# Improved ICNN-LSTM Model Classification Based on Attitude Sensor Data for Hazardous State Assessment of Magnetic Adhesion Climbing Wall Robots

Zhen Ma[a], He Xu[a],*, Jielong Dou[a], Yi Qin[a], Xueyu Zhang[a]
[a]College of Mechanical and Electrical Engineering, Harbin Engineering University, Harbin 150001, China. No.145, Nan Tong Street, Nan Gang District.

**Abstract:** Magnetic adhesion tracked climbing robots are widely utilized in high-altitude inspection, welding, and cleaning tasks due to their ability to perform various operations against gravity on vertical or inclined walls. However, during operation, the robot may experience overturning torque caused by its own weight and load, which can lead to the detachment of magnetic plates and subsequently pose safety risks. This paper proposes an improved ICNN-LSTM network classification method based on Micro-Electro-Mechanical Systems (MEMS) attitude sensor data for real-time monitoring and assessment of hazardous states in magnetic adhesion tracked climbing robots. Firstly, a data acquisition strategy for attitude sensors capable of capturing minute vibrations is designed. Secondly, a feature extraction and classification model combining an Improved Convolutional Neural Network (ICNN) with a Long Short-Term Memory (LSTM) network is proposed. Experimental validation demonstrates that the proposed minute vibration sensing method achieves significant results, and the proposed classification model consistently exhibits high accuracy compared to other models. The research findings provide effective technical support for the safe operation of climbing robots.

**Keywords:** Magnetic Adhesion Climbing Robot, MEMS Attitude Sensor, Long Short-Term Memory Network, Hazard Assessment, Deep Learning

1. Introduction

Magnetic adhesion tracked climbing robots are specialized machines capable of performing various tasks against gravity on vertical or inclined walls [1]. Consequently, these climbing robots are well-suited for high-altitude work environments such as inspection [2], welding [3], and cleaning [4][5], thereby ensuring the safety of operators [6]. However, when moving on vertical or inclined surfaces, the robot's own weight and the load it carries can generate overturning torque, potentially causing the robot body to tilt backward [7]. Therefore, ensuring sufficient magnetic plate adhesion and establishing effective contact between the magnets and the wall surface [8-9] are crucial for the safe operation of climbing robots. To address this issue, it is necessary to develop an effective perception method to monitor the magnetic adhesion state during the robot's movement [10]. Magnetic climbing robots typically rely on magnetic adhesion to climb and secure themselves on ferromagnetic surfaces [11-12]. Tactile sensors placed between the robot's magnetic plates and the ferromagnetic surface can either facilitate or hinder direct contact between the robot and the wall, thereby affecting the stability and reliability of the magnetic adhesion [13].

MEMS sensors are not commonly used for fault diagnosis; however, they play an important role in robot balance control, human motion analysis, and aircraft attitude measurement. As a type of inertial measurement unit, attitude sensors are widely applied in mobile robots, drones, and other fields to monitor the robot's motion state and posture changes in real time [14-16]. MEMS sensors can collect data signals such as angular velocity, vibration acceleration, and magnetic field strength, which are critical for analyzing and assessing the robot's operational state [17]. However, as low-cost commercial sensors, attitude sensors exhibit characteristics like significant drift and low accuracy. Enhancing their sensitivity to detect subtle state changes in the robot without losing the inherent features of the vibration signals is both meaningful and necessary [18].

With advancements in computational technology, traditional machine learning methods have achieved certain successes in robot fault diagnosis. However, these methods often suffer from long processing times, low accuracy, and a dependence on manual feature selection [19-20]. Deep learning approaches, particularly hybrid models that combine Convolutional Neural Networks (CNN) and Long Short-Term Memory (LSTM) networks, have emerged as important solutions to these challenges due to their powerful feature extraction and time-series modeling capabilities [21-23].

The main contributions of this paper are twofold: (1) Designing a data acquisition strategy based on MEMS sensors for monitoring hazardous states in magnetic adhesion tracked climbing robots; and (2) Proposing a feature extraction and classification model that integrates an Improved Convolutional Neural Network (ICNN) with an LSTM network for the discrimination of hazardous states in climbing robots. The structure of this paper is organized as follows: Section 2 provides a review of existing research. Section 3 discusses the analysis of the robot overturning risk mechanism, the data acquisition strategy, the proposed ICNN-LSTM network, and the risk classification procedure. Section 4 presents the preparation of experimental data and the experimental results. Finally, Section 5 concludes the paper.。

2. Literature Review

2.1 Attachment State Perception Technology for Magnetic Adhesion Climbing Robots

In previous studies, pressure sensors have been commonly used to monitor and detect the attachment state of climbing robots [24-25]. However, the magnetic adhesion force significantly decreases as the distance between the magnet and the ferromagnetic metal wall surface increases, making pressure sensors unsuitable for installation on the magnetic adhesion units of magnetic climbing robots [26]. Vacuum sensors and photoelectric sensors have also been employed to ensure the stable and reliable adhesion of wall-climbing robots to climbing surfaces [27]. However, vacuum sensors are not applicable to magnetic adhesion climbing robots, and photoelectric sensors have failed to detect the reliability and stability of magnetic adhesion. Vision sensors have also been utilized [28], but these sensors perform inadequately in dark lighting conditions and environments with multiple obstacles. Existing research generally requires the installation of complex and expensive sensor systems on robots, with limited studies on convenient and low-cost monitoring and hazard assessment of magnetic adhesion states. Specifically for magnetic adhesion climbing robots, there are currently no mature and reliable solutions for detecting the attachment state of magnetic plates.

2.2 Applications of Micro-Electro-Mechanical Systems (MEMS) in Mechanical-Related Fields and Sensitivity Amplification Methods

Micro-Electro-Mechanical Systems (MEMS) data are utilized by pipeline robots to assess the robot's position and orientation, effectively enhancing the robot's stability and safety [29]. MEMS data are also used to monitor the operational status of 3D printers, achieving health condition diagnostics for 13 types of faults, thereby improving the maintainability and stability of faulty printers [30]. In the automotive field, MEMS data are employed for precise vehicle yaw estimation and dead reckoning, enabling the implementation of low-cost driver assistance systems [31].

Regarding sensitivity amplification methods for MEMS sensors, micro-mechanical lever systems are used to enhance deflection inputs caused by inertial forces [32]. Hydraulic displacement amplification mechanisms have also been employed for MEMS amplification, consisting of fully encapsulated fluids and deformable membranes [33]. Planar compliant amplifier mechanisms have been applied to amplify the sensitivity of MEMS sensors [34]. Various methods such as bridge structures, positioning stage amplifiers, Scott-Russell mechanisms, multi-stage force-displacement amplifiers, and thermally driven displacement amplifiers have achieved certain effects in enhancing the sensitivity of MEMS sensors [35]. However, these methods generally involve complex mechanical designs, high manufacturing and maintenance costs, making them unsuitable for non-precision applications like climbing robots. Although MEMS sensors have been effectively applied in numerous fields and their sensitivity can be amplified, their application in the perception of attachment states in climbing robots is rare.

2.3 Applications of CNN-LSTM in Fault Diagnosis and Related Fields

In recent studies, the combination of Convolutional Neural Networks (CNN) and Long Short-Term Memory (LSTM) networks has demonstrated outstanding performance in the fields of fault diagnosis and condition monitoring [36]. In the field of mechanical fault diagnosis, CNN-LSTM networks have been used for fault classification of vibration data from helicopter gearbox systems, achieving defect identification of gear root cracks [37]. CNN-LSTM has also been applied to bearing fault diagnosis, showing better performance in noisy environments [38]. Similarly, CNN-LSTM has achieved good results in fault diagnosis of hydraulic systems, with a fault diagnosis accuracy rate reaching 98.56%. The CNN-LSTM model has proven to be an effective mechanical fault diagnosis model [39-40]. However, there is still a lack of systematic deep learning method research for the hazard state assessment of magnetic adhesion climbing robots.

3. Methodology

This study proposes a hazard assessment system for climbing robots utilizing carbon fiber vibration rod sensors and an improved ICNN-LSTM model. The proposed framework consists of two main components: a sensor sensitivity amplification method based on carbon fiber vibration rods and a hazardous state classification method based on the improved CNN-LSTM model.

The first component aims to provide a low-cost and straightforward approach for detecting subtle vibrations in climbing robots. The second component employs the improved

ICNN-LSTM model to identify hazardous states of the climbing robot, specifically the number of magnetic plate attachments on the robot's tracks. Notably, this component includes three steps:1)Data Preprocessing: This step involves removing high-frequency noise from the signals to obtain high-quality vibration data.2)Feature Extraction: An improved deep neural network (ICNN) architecture is utilized to extract meaningful features from the selected attitude sensor signals.3)Hazardous State Classification: A classification model based on Long Short-Term Memory (LSTM) networks is developed to determine the number of magnetic plate attachments on the robot's tracks. Figure 1 provides an overview of the proposed framework. Further details regarding the different steps of the framework will be explained in the subsequent sections。

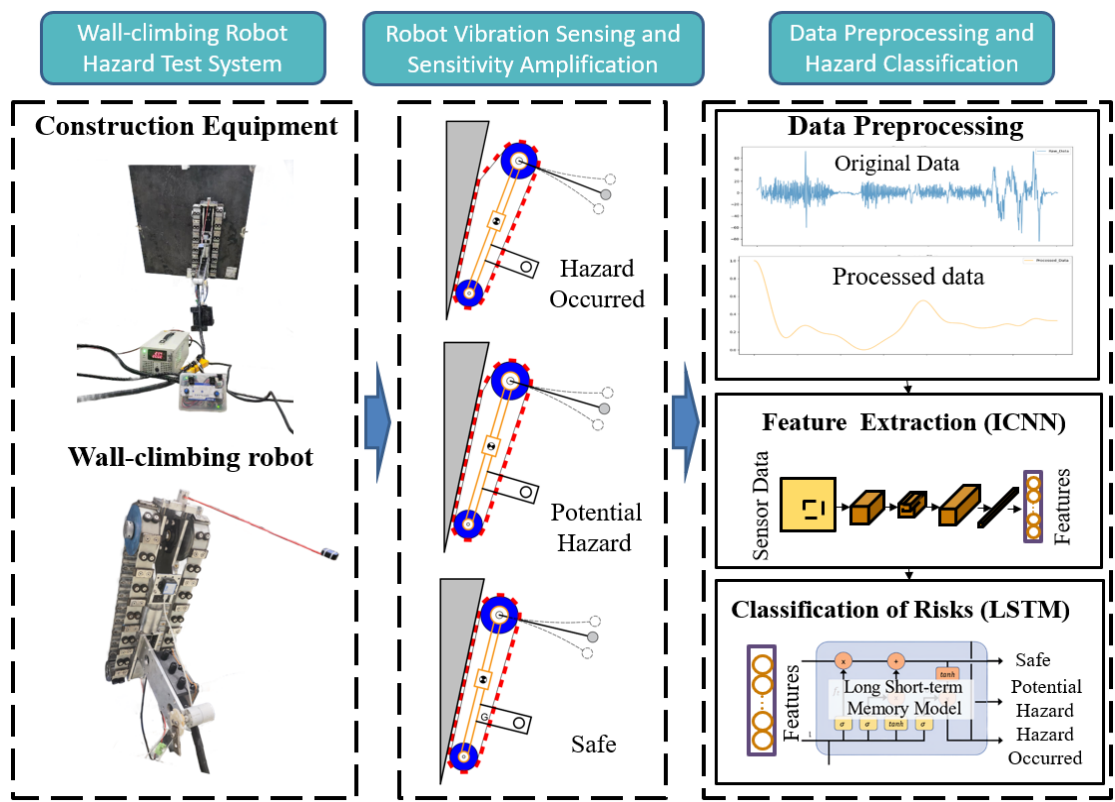

Fig. 1. Overview of the proposed framework

3.1 Attachment Mechanism and Adhesion Stiffness Analysis of Magnetic Plates for Climbing Robots

3.1.1 Attachment Mechanism of Magnetic Plates on Robot Tracks

Under conditions of high load or steep wall inclinations, magnetic adhesion-based tracked climbing robots may experience gradual detachment of their tracks from the wall surface, as illustrated in Figure 2. During the upward climbing process, the robot tracks deform due to the combined effects of the load and the wall's inclination angle. It can be observed that when the magnetic plates roll into the front adhesion area with the tracks, the angle θ increases, resulting in an increase in the distance $L_h$ between the magnets and the wall surface. This leads to a significant weakening of the magnetic attraction, and the restorative force generated by the track deformation becomes insufficient to maintain normal adhesion. To facilitate a clearer and more straightforward analysis, the examination of the magnetic plate lifting in climbing robots is transformed into an analysis of the magnetic plate's adhesive force.

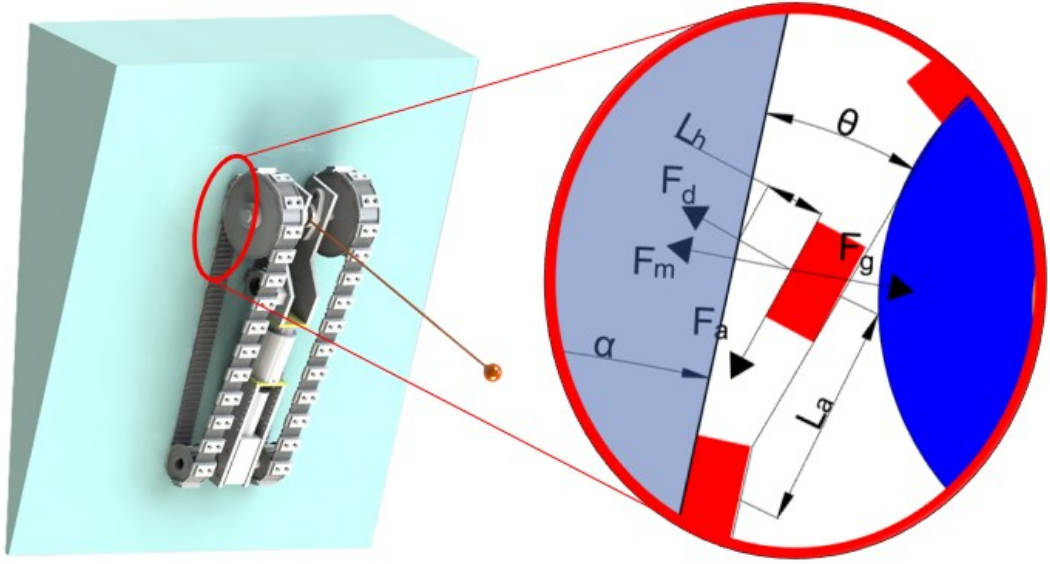

Fig.2. Force analysis of magnetic disk of wall-climbing robot

When the robot climbs upward, as the tracks roll into the front adhesion area, the magnetic attraction force $F_m$ of the magnetic plates comprises the sum of magnetic force, track restorative force, and tension force components. In contrast, Fh represents the opposing resultant force caused by gravity and load. To simplify the description of the interaction between these forces, a matrix representation is used.

$$F_m = \begin{bmatrix} \dfrac{k}{L_h^2} \\ F_d \cos\theta \\ F_a \cos\theta \end{bmatrix}$$

$$F_g = \left[(G_a + G_b)\sin(\theta + \alpha)\right]$$

Here, $k$ is the magnetic force coefficient, $L_h$ is the distance from the magnetic plate to the wall surface, $F_d$ and $F_a$ are the track's restorative and tension forces, respectively, $\theta$ is the track's bending angle, $G_a$ and $G_b$ are the gravitational forces of the robot and the load, and $\alpha$ is the wall's inclination angle. It is evident that when $||F_m|| \geq ||F_g||$, the magnetic plates adhere to the wall; when $||F_m|| < ||F_g||$, the magnetic plates fail to adhere. In cases where adhesion fails, the robot experiences a gradual reduction in the number of adhered magnetic plates, ultimately leading to a complete detachment.

3.1.2 Adhesion Stiffness Analysis

The connection between the climbing robot and the wall surface is collectively formed by the rigid connections of each magnetic plate. Let us assume that the number of magnetic plates attached by the magnetic adhesion tracked robot at a certain moment is $N$. Then, the total contact stiffness $k$, the system's natural frequency $\omega_n$, and the damping ratio $\zeta$ are defined as follows::

$$k = N \cdot k_i$$

$$\omega_n = \sqrt{\frac{k}{m}} = \sqrt{\frac{N \cdot k_i}{m}}$$

$$\zeta = \frac{c}{2\sqrt{mk}} = \frac{c}{2\sqrt{mNk_i}}$$

From the above equations, it is evident that as $N$ increases or decreases, the robot's total contact stiffness $k$ correspondingly increases or decreases. This leads to an increase or decrease in the robot's natural frequency $\omega_n$, causing the system's vibration to become faster or slower. Additionally, the damping ratio $\zeta$ changes accordingly. Consequently, when the number of attached magnetic plates varies, the climbing robot exhibits different responses to vibrations caused by itself or the environment.

3.2 Analysis of the Sensitivity Amplification Mechanism for MEMS Attitude Sensors Based on Carbon Fiber Rods

MEMS attitude sensors are connected to the robot body via carbon fiber rods, which can amplify the vibration states of the robot. However, the introduction of carbon fiber rods significantly impacts both the signal synchronization and the amplification effect. Through systematic dynamic modeling and frequency response analysis, it can be demonstrated that the sensitivity amplification mechanism introduced by the fiber rods achieves both signal synchronization and amplification of the sensor signals.。

Here, we assume that the robot's state equation is $x_1(t)$, and the state equation at the top of the fiber rod is $x_2(t)$. Let $k$ be the stiffness of the fiber rod and $c$ be the damping coefficient, while neglecting the mass of the fiber rod. The force $F$ exerted on the top of the fiber rod can be expressed as:

$$F = k(x_1(t) - x_2(t)) + c(\dot{x}_1(t) - \dot{x}_2(t))$$

Convert the equation of motion to the frequency domain:

$$m \cdot \omega^2 X_2(\omega) = k(X_1(\omega) - X_2(\omega)) + j\omega c(X_1(\omega) - X_2(\omega))$$

The resulting transfer function is: $H(\omega) = \dfrac{X_1(\omega)}{X_2(\omega)} = \dfrac{k + j\omega c}{k - M\omega^2 + j\omega c}$

Calculate the gain and phase over the target frequency range:

$$|H(\omega)| = \frac{\sqrt{k^2 + (\omega c)^2}}{\sqrt{(k - M\omega^2)^2 + (\omega c)^2}}$$

$$\phi(\omega) = tan^{-1}(\frac{\omega c}{k}k) - tan^{-1}(\frac{\omega c}{k - M\omega^2})$$

It is evident that when the stiffness $k$ decreases, the gain $|H(\omega)|$ increases, resulting in the amplification of the vibration signals from the robot body and an improvement in sensing sensitivity. However, this also leads to the enhancement of low-frequency vibration signals, thereby distorting the signal frequency characteristics. Conversely, when the stiffness $k$ increases, the original signal characteristics are better preserved, but the amplification effect diminishes, leading to a decrease in sensor sensitivity.

Based on preliminary experiments, we have determined the use of carbon fiber rods with a diameter of 3 mm and a length of 500 mm for the robots in this study. This specification of carbon fiber rods balances the requirements for signal amplification and the preservation of frequency domain characteristics in the transmission of vibration signals from the robot。

3.3 Improvements to the CNN-LSTM Network and Application in Safety State Classification

3.3.1 Overall Structure of the Proposed Framework

To utilize the robot's vibration signals (i.e., signals from the MEMS attitude sensors) for discriminating hazardous states of the climbing robot, the authors developed a classification framework based on the improved ICNN-LSTM model, as illustrated in Figure 3. The process consists of the following steps: 1) Data Preprocessing: The input vibration signals undergo preprocessing to eliminate the influence of high-frequency noise. 2) Feature Extraction: The proposed Adaptive Convolutional Neural Network (ICNN) framework is applied to extract informative features from the robot's vibration signals. 3) Safety State Classification: A Long Short-Term Memory (LSTM) model is employed to classify the extracted features, determining the hazardous state of the climbing robot. Each component of the framework is elaborated in the subsequent subsections.

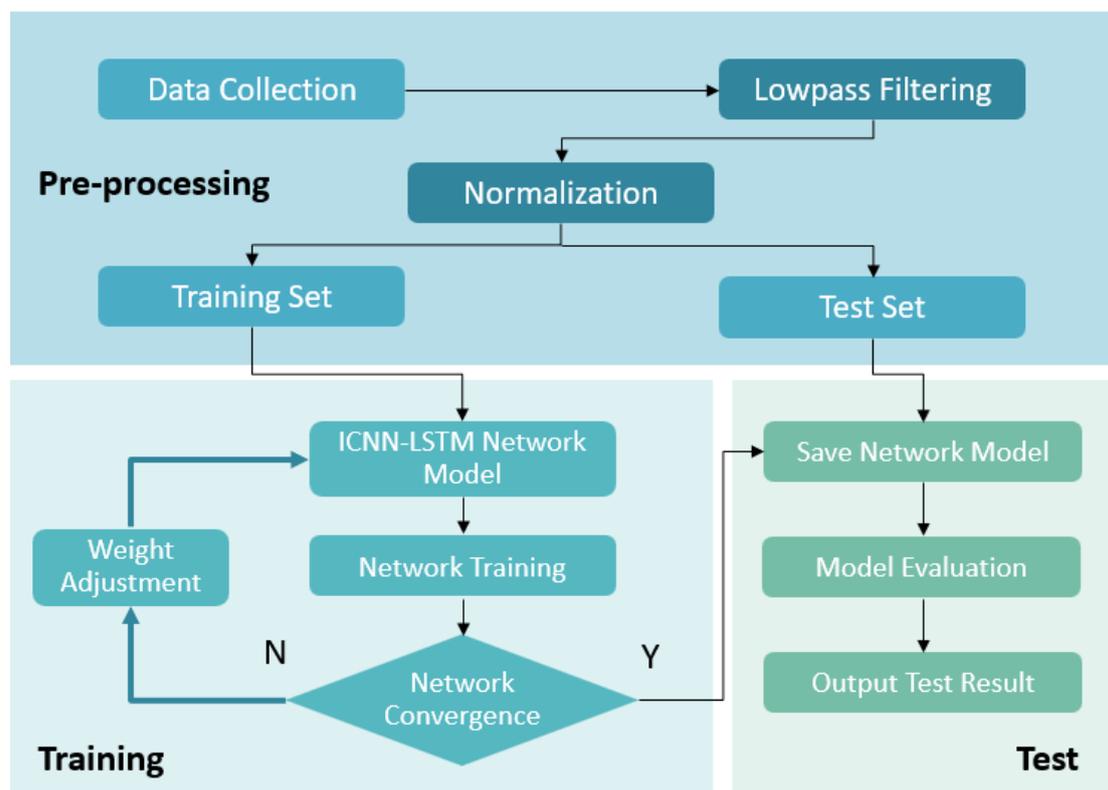

Fig. 3. Wall climbing robot danger state classification flow chart.

3.3.2 Data Preprocessing

As discussed in Section 3.1, the climbing robot experiences changes in system vibration characteristics due to variations in the number of attached magnetic plates, which typically occur in the low-frequency domain. To effectively extract low-frequency signals from the MEMS attitude sensors and suppress high-frequency noise, a Low-Pass Filter (LPF) was employed for signal processing. The specific implementation process includes two main steps:

filtering and normalization, as detailed below:

In this study, a Butterworth Filter was selected due to its maximally flat amplitude frequency response, which is suitable for applications requiring smooth frequency responses. The design of the filter is based on the following mathematical model:

$$H(s) = \frac{b_0 + b_1 s + \cdots}{a_0 + a_1 s + \cdots}$$

where $H(s)$ is the transfer function of the filter, $b_i$ and $a_i$ are the filter coefficients, and $n$ is the order of the filter.

After filtering, the MinMaxScaler was used to normalize the filtered data to the range [0,1], according to the following formula:

$$x' = \frac{x - x_{min}}{x_{max} - x_{min}}$$

where $x$ is the original data, $x_{min}$ and $x_{max}$ are the minimum and maximum values of the data, respectively, and $x'$ is the normalized data.

It is noteworthy that after calibration, such sensors still display a gravitational acceleration of 1g. Additionally, inconsistencies in the initial installation orientations of the sensors lead to increased variability in the training data, thereby reducing the model's generalization performance. To eliminate the impact of variations in sensor installation orientation and gravity direction on model training and prediction performance, this study computes the magnitude of the three-axis acceleration vector output by the sensor. By calculating the magnitude, the data is transformed into scalar values as input features, thus avoiding data errors caused by inconsistent installation orientations. The resulting input signal can be represented as:

$$F_{input} = \| \quad \| \quad \overline{a_{yi}^2 + a_{zi}^2}$$

where $\|a_i\|$ is the magnitude of the acceleration calculated by the sensor at the $i$-th moment, and $F_{input}$ is the input feature matrix for the model.

3.3.3 Hazard State Classification: Long Short-Term Memory Layer (LSTM Layer)

In constructing the model, the Improved Convolutional Neural Network (ICNN) architecture integrates classical convolution operations with an adaptive ReLU activation function to enhance the network's expressive capability and classification performance. This structure comprises multiple convolutional blocks, each containing a one-dimensional convolutional layer, an adaptive activation layer, a max-pooling layer, and a Dropout layer. These components provide rich information for further discriminative feature extraction.

In the convolutional layers, we utilize classical fixed convolution kernel sizes and a set number of filters. Specifically, a convolution kernel size of 3 and 64 filters are employed to extract local features from the input data, effectively capturing short-term dependencies within the sequence data. Following each convolutional block, an adaptive ReLU activation

layer is added to further enhance the network's nonlinear expressive capabilities, defined as follows::

$$p_{li}(t) = max(0, z_{li}(t)) + \alpha_{li} min(0, z_{li}(t))$$

where $p_{li}(t)$ represents the output of the $i$-th neuron in the $l$-th layer after the adaptive ReLU activation, $z_{li}(t)$ denotes the input feature value to the Batch Normalization (BN) layer, and $\alpha_{li}$ is the adaptive parameter to be learned for that layer. The design of the adaptive ReLU aims to enhance the network's ability to handle negative values by utilizing an adaptive negative slope, thereby increasing the model's capability to adjust the feature distribution while maintaining nonlinear expression.

Additionally, to prevent feature redundancy and overfitting, each convolutional block includes a max-pooling layer and a Dropout layer. The max-pooling layer uses a pooling window size of 1 to reduce feature dimensionality and enhance the selectivity of local features. (Note: A pooling window size of 1 typically does not perform any pooling operation. It is possible that this was intended to be a different size, such as 2 or 3. Please verify this parameter for accuracy.) The Dropout layer discards 20% of the neurons within each convolutional block, further reducing the risk of overfitting.

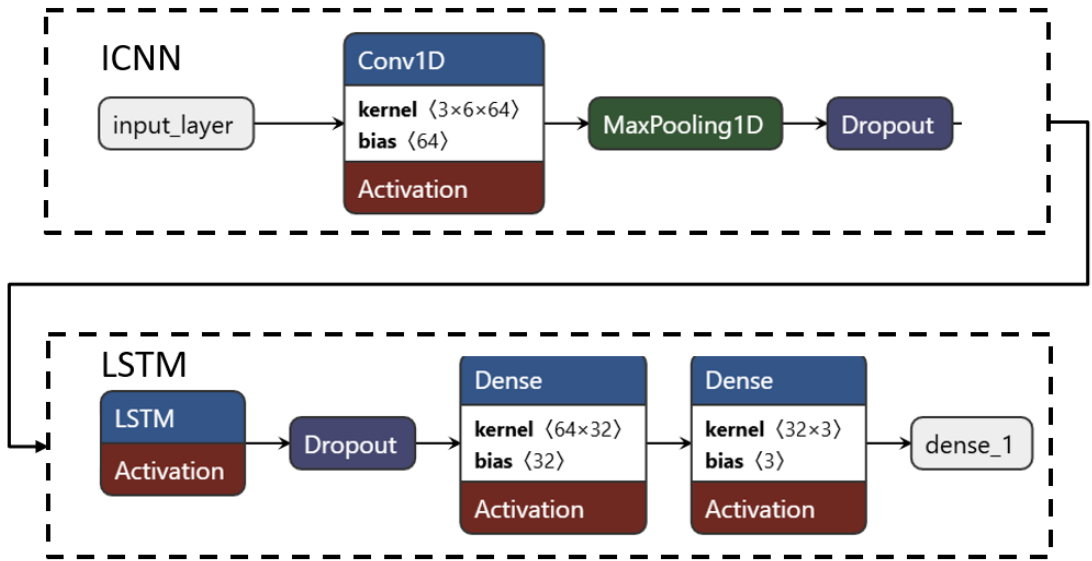

Fig. 4. Network Structure of ICNN-LSTM

3.3.3 Hazard State Classification: Long Short-Term Memory Layer (LSTM Layer)

In the Integrated Convolutional Neural Network (ICNN) combined with Long Short-Term Memory (LSTM) network (ICNN-LSTM), the LSTM layer is designed to extract temporal features and perform further temporal modeling of the local features encoded in the convolutional feature maps. The LSTM layer comprises multiple memory cells that regulate information updates at each time step through input gates, forget gates, and output gates. Initially, the output from the convolutional blocks is fed into the LSTM layer, where the LSTM units sequentially process information at each time step to form a comprehensive representation of temporal dependencies.

The output $f_t$ of each LSTM unit is generated by the combination of the internal memory cell $c_t$ and the output gate $o_t$. The update process of the memory cell $c_t$ is as follows:

$$c_t = f_t \odot \quad \odot$$

where $f_t$, $i_t$, and $\tilde{c}_t$ represent the forget gate, input gate, and candidate memory cell state, respectively. The forget gate determines the extent to which past information is retained or forgotten, the input gate decides the degree to which current information is stored, and the candidate memory cell state temporarily holds new information. Finally, the output of the LSTM unit $f_t$ is generated through the output gate $o_t$, calculated as follows::

$$h_t = o_t \odot$$

Through this gating mechanism, the LSTM can dynamically update its state at each time step, thereby capturing long-term dependencies in the temporal data.

Finally, all outputs of the LSTM layer are passed through a flatten operation to generate a final discriminative feature map $F$. This feature map can be further input into a classifier consisting of fully connected layers and a Softmax layer to obtain classification probability outputs for the time series data. Additionally, the feature map can also be utilized for cross-domain feature extraction within a transfer learning framework to achieve stronger generalization capabilities under different data conditions.。

3.4 Steps for Hazard State Classification of Climbing Robots

The ICNN network is integrated with the LSTM network to form the ICNN-LSTM network, as shown in Figure 4. Algorithm 1 outlines the implementation process of the ICNN-LSTM classification algorithm.

**Algorithm:** ICNN-LSTM Model Training and Evaluation

**Input**: Dataset D = { (x₁, y₁), (x₂, y₂), ..., (x_N, y_N) }

1: Data Preprocessing; Normalization parameters (mean μ, standard deviation σ)
    For each sample x_i ∈ D:
        x_i_normalized = (x_i - μ) / σ

2: Dataset Splitting; Shuffle the dataset D randomly; Training set ratio α
    Split into training and testing sets:
        Training set size = α * N
        Training set D_train = { (x₁, y₁), ..., (x_{αN}, y_{αN}) }
        Test set D_test = { (x_{αN+1}, y_{αN+1}), ..., (x_N, y_N) }

3: Model Training
    Initialize ICNN-LSTM model parameters
    Repeat until convergence or reaching maxSensorm iterations:
        For each batch B ⊂ D_train:
            Feature Extraction:
                C = ConvolutionLayer(x_B_normalized)
            Sequence Modeling:
                L = LSTM_Layer(C)
            Classification Prediction:
                logits = FullyConnectedLayer(L)
                y_pred = Softmax(logits)
            loss = CrossEntropy(y_pred, y_B)
            Backpropagate(loss)

```
              Update ICNN-LSTM parameters
    4: Model Evaluation
        Initialize correct_predictions = 0
        For each sample (x, y) ∈ D_test:
              x_normalized = (x - μ) / σ
              C = ConvolutionLayer(x_normalized)
              L = LSTM_Layer(C)
              logits = FullyConnectedLayer(L)
              y_pred = Softmax(logits)
              y_label = ArgMax(y_pred)
              If y_label == y:
                    correct_predictions = correct_predictions + 1
              accuracy = correct_predictions / |D_test|
```
**Output:** Prediction accuracy on the test set, y_pred

## 4. Experiments

### 4.1 Experiment Setup

To evaluate the effectiveness of the proposed magnetic adhesion tracked climbing robot's hazard perception and classification, a robot drop test system was established for testing and data collection. As shown in Figure 5, the system comprises an operator, a tracked robot, body-attitude sensors, fiber-rod-attitude sensors, an angle-adjustable steel plate, and a load module.

Experiments were conducted in an indoor laboratory environment, where the steel plate was securely fixed to the frame, and the operator, who had undergone training, operated the system proficiently and stably. The climbing robot is designed to support a load of 5 kg; therefore, the robot was consistently loaded with a fixed 5 kg. In this experiment, the only variable was the angle of the climbing steel plate. When the robot was loaded with 5 kg, the angle between the adjustable steel plate and the vertical plane was set to two angles: 55° and 65°. These two angles represent the climbing robot's two extreme working angles.。

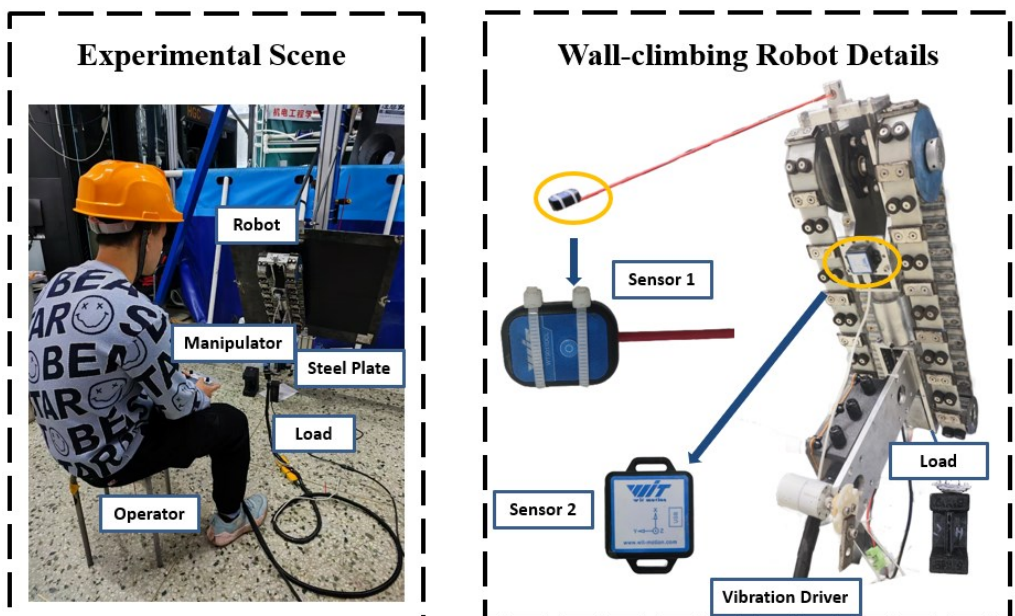

Fig. 5. Wall-climbing robot drop test system

### 4.2 Composition and Installation of the Robot and Sensors

The magnetic adhesion tracked climbing robot consists of two tracks and magnetic plates. Each track is equipped with a permanent magnet plate measuring 20 × 40 mm, installed every 80 mm. Each magnetic plate can provide a magnetic adhesion force of 200 N. One end of the carbon fiber rod is rigidly connected to the robot body, while the other end is fixed to an attitude sensor. Additionally, another attitude angle sensor is installed on the robot body to compare the sensing performance of the carbon fiber rod. The sensor on the fiber rod is connected via 5G communication, whereas the sensor on the robot body is connected through USB communication. Both sensors are three-axis attitude angle measurement sensors with identical parameters and precision, as detailed below:

- Measurement Range: ±2000 °/s
- Sampling Frequency: 0.2–100 kHz
- Resolution: 0.061 (°/s)/(LSB)
- Static Zero Bias: ±0.5~1 °/s
- Temperature Drift: ±0.005~0.015 (°/s)/°C
- Sensitivity: ≤0.015 °/s RMS

The sampling frequency of both attitude sensors is set to 100 Hz. When the robot moves along a predefined trajectory, nine-channel signals are collected from the attitude sensors, all of which are timestamped for convenient temporal alignment. The robot's forward speed is 0.01 meters/second. Data collection begins with all six magnets fully adhered. The experiment stops when only four magnets remain adhered. To ensure the accuracy of the labels, data corresponding to partial separation or partial contact of the magnets were not collected. After completing one run, 20,000 sample points can be collected. Each angle measurement is repeated twice, resulting in a total of 60,000 sample points. Of these, 70% of the data is used for training, and 30% is used for testing.

Table 1. Hazardous States of the Climbing Robot and Corresponding Labels

| Serial Number | Actual State | Label |
| --- | --- | --- |
| 1 | 6 Magnetic Plates Attached | Safe |
| 2 | 5 Magnetic Plates Attached | Potential Hazard |
| 3 | 4 Magnetic Plates Attached | Hazard Occurred |

As introduced in Section 2.1, during the climbing process, the initial state of the climbing robot involves all six magnetic plates being fully adhered. The robot begins to climb the wall, collecting data and labeling accordingly. When five magnetic plates remain adhered, it is considered that a potential hazard has emerged, and data are collected and labeled accordingly. The robot continues to operate, and when only four magnetic plates remain adhered, it is deemed that a hazard has occurred, and data are collected and labeled accordingly. Table 1 illustrates the relationship between the actual states of the robot and their corresponding labels. After only four magnetic plates remain adhered, the robot returns to the initial position to restart the experiment.

Experiments were conducted twice on steel plates inclined at 55° and 65°, respectively. Each repeated experiment was further divided into four groups with vibration excitation intensities ranging from level 0 to level 4. The angular velocities along the XYZ axes and the corresponding state labels were recorded to construct a comprehensive dataset for training the hazard classification model, enabling the identification of specific hazard categories.。

## 5. Experimental Results and Discussion

### 5.1 Basic Parameter Settings

Based on the experimental design using MEMS attitude sensor data from the climbing robot, the basic parameter settings for the proposed ICNN-LSTM model are as follows: the Adam optimizer was employed with a learning rate set to 0.001. During model training, an early stopping strategy (patience=5) and a model checkpoint strategy were used to save the best model parameters. The model was trained for 30 epochs with a batch size of 32. A one-hot encoding method was adopted for label classification, and the model performance was optimized using the categorical cross-entropy loss function. All code was implemented using the TensorFlow and Keras frameworks and executed on a computing platform equipped with an NVIDIA 3060 12GB GPU.

### 5.2 Data Quality Analysis of the Two Attitude Sensors

To investigate the performance of the proposed vibration rod capable of capturing subtle vibrations, an adjustable eccentric wheel motor was used as the vibration exciter with excitation intensities set to four levels: 0, 1, 2, and 3.

First, we compared the data quality of signals from the two sensors under different vibration excitation levels. Based on the concepts of data quality proposed in [41-43], we evaluated the signal quality from multiple dimensions, including multidimensionality of data quality, dynamic characteristics, anomaly detection and error assessment, and quantitative signal characteristics. Specifically, we established six dimensions for evaluating data signal quality: signal energy, signal standard deviation (STD), kurtosis, skewness, power spectral density (PSD), and spectral centroid. Feature extraction for the original data was conducted as shown in Table 2:

Table 2. Comparative Evaluation of Sensor Data Signal Quality

| Metric | vibrational_0 | vibrational_1 | vibrational_2 | vibrational_3 |
|---|---|---|---|---|
| Energy_1 | 62123.16681 | 61480.82611 | 58988.95717 | 61955.06486 |
| Energy_2 | 157818.6077 | 160238.4459 | 150552.3913 | 162516.789 |
| STD_1 | 0.014678165 | 0.015164309 | 0.016588302 | 0.017063531 |
| STD_2 | 0.005317324 | 0.016448709 | 0.027525445 | 0.043312097 |
| Kurtosis_1 | -1.454811132 | -1.374345505 | -0.865131844 | -0.70365 |
| Kurtosis_2 | 4.560958 | 1.473049 | 0.354874 | -0.19825 |
| Skewness_1 | -0.34713 | -0.26935 | -0.24156 | -0.25382 |
| Skewness_2 | -0.77669 | 0.032744 | 0.02302 | 0.022652 |
| Spectral_Centroid_1 | 47.23476 | 117.0473 | 157.4615 | 127.551 |
| Spectral_Centroid_2 | 27.84928 | 115.6345 | 127.9449 | 76.13154 |

Under all excitation intensities, SENSOR_2 exhibited significantly higher signal energy compared to SENSOR_1, indicating that SENSOR_2 is capable of capturing stronger vibration signals. Additionally, the signal standard deviation of SENSOR_1 was consistently lower than that of SENSOR_2 across all excitation levels, especially under high excitation intensities, where SENSOR_1 demonstrated smaller signal fluctuations and higher signal stability.

From the kurtosis analysis, SENSOR_1 showed negative kurtosis under all excitation intensities, indicating a flatter signal distribution with fewer peaks and outliers. In contrast, SENSOR_2 exhibited higher kurtosis at low to medium excitation levels, suggesting the presence of more peaks or noise in the signals. However, at high excitation intensities,

SENSOR_2's kurtosis became negative, indicating a flatter signal distribution. This phenomenon, as described in Section 3.2, is due to the vibration transmission characteristics of the slender rod, which lead to the degradation of high-frequency features.

Regarding skewness, SENSOR_1 consistently displayed negative skewness across all excitation intensities, particularly showing larger negative skewness at excitation level 0, indicating a left-skewed signal characteristic. In comparison, SENSOR_2's skewness was close to zero for most excitation levels, except at excitation level 0, where it exhibited significantly negative skewness. This variation is primarily caused by differences in sensor installation and robot operating conditions, resulting in no significant advantage between the two sensors in terms of skewness.

For the spectral centroid, SENSOR_1 had higher spectral centroids across all excitation intensities compared to SENSOR_2, especially at excitation levels 1 to 3, indicating that SENSOR_1 can capture more high-frequency signals. SENSOR_2, however, showed lower spectral centroids at excitation levels 0 and 1, although there was an increase at excitation levels 2 and 3, it remained generally lower than SENSOR_1.

Overall, under low to medium excitation intensities (0 to 2), SENSOR_2 provided better signal quality than SENSOR_1, making it suitable for capturing stronger vibration signals, albeit with some loss of high-frequency information. At high excitation intensity (vibrational_excitation_3), SENSOR_1 outperformed SENSOR_2 in signal quality.

To further investigate the data from the two sensors, we applied the proposed ICNN-LSTM model for comparative classification. During training, all model parameters were kept consistent, with only the dataset being varied to evaluate training outcomes, as shown in Figure 6.

It is evident that the two sensors, SENSOR_1 and SENSOR_2, exhibit significant differences under various excitation conditions. When excitation levels range from 0 to 2, classification training results using SENSOR_2 data are significantly better than those using SENSOR_1 data. SENSOR_2 achieved optimal classification performance at excitation level 2. However, when the excitation level increased to 4, the classification results of SENSOR_1 surpassed those of SENSOR_2.

Combining the discussion in this section, it is understood that the presence of the vibration rod indeed amplifies the robot's vibrations. However, when the robot body experiences sufficiently large vibrations, the amplification effect of the vibration rod can have adverse impacts. This is consistently reflected in both the data quality analysis and the model training comparisons.

5.3 Comparison of Training Results Among Different Models

To demonstrate the superiority of the proposed ICNN-LSTM classification model, we conducted comparative analyses using data from both sensors. Given that the excitation intensity level 3 closely resembles the environmental conditions applied in this study, SENSOR_2 data at excitation level 3 was utilized during the model training comparison phase. The ICNN-LSTM model was compared against six classical methods, including standard LSTM, Recurrent Neural Network (RNN), Backpropagation Neural Network (BP), Random Forest (RF), and K-Nearest Neighbors (KNN). Below is a brief introduction to the four primary models compared:

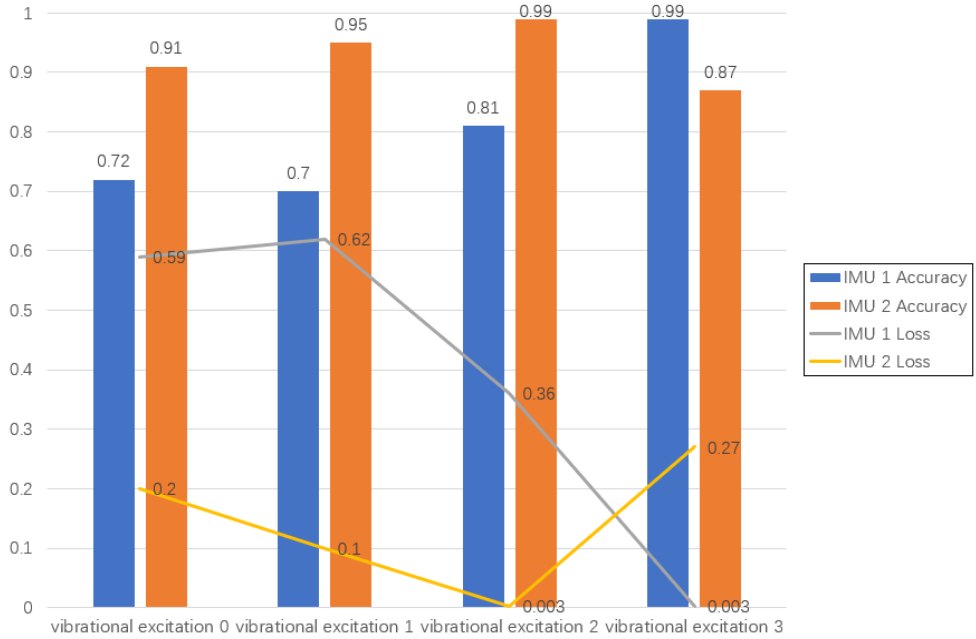

Fig. 6. Comparison of training accuracy and loss of two sensors under different excitation conditions

Long Short-Term Memory Network (LSTM): LSTM is a variant of Recurrent Neural Networks (RNNs) capable of effectively capturing long-term dependencies in time-series data. By introducing forget gates, input gates, and output gates, LSTM overcomes the limitations of traditional RNNs, such as vanishing or exploding gradients. This study employed a standard LSTM network comprising three LSTM layers and two fully connected layers, which models data dependencies over time steps to perform classification tasks, as implemented based on reference [45].

Recurrent Neural Network (RNN): RNNs are widely used for processing sequential data through their recursive connection structure, enabling the learning of dependencies between time steps in sequence data. This study utilized a standard RNN model with two recurrent layers and two fully connected layers to extract and classify sequential features. To further optimize model performance, a truncated backpropagation through time technique was employed to mitigate the vanishing gradient problem, following the model architecture outlined in reference [46].

Backpropagation Neural Network (BP): BP is a classical shallow learning method centered on adjusting weights and biases through the error backpropagation algorithm to optimize the objective function. This study constructed a three-layer BP neural network model, where the input layer handles feature inputs, the hidden layers consist of two non-linear activation layers each with 100 nodes, and the output layer is designed for multi-class classification. The model architecture was referenced from .

Random Forest (RF): RF is an ensemble learning method that constructs multiple decision trees and employs a majority voting mechanism to achieve classification tasks. RF builds each decision tree by randomly selecting samples and features, which provides robust resistance to overfitting. This study built an RF model with 100 decision trees using the Gini index as the splitting criterion and optimized hyperparameters through cross-validation, as detailed in reference .

K-Nearest Neighbors (KNN): KNN is an instance-based, non-parametric classification method that assigns class labels based on the majority vote of the K closest training samples, using Euclidean distance as the similarity measure. This study implemented a KNN model with K=5 to efficiently classify multi-class data, following the implementation approach described in reference .

All methods were trained five times independently using the same sample dataset to verify the stability of the model results. A unified testing dataset was used for performance evaluation. The experimental results, as depicted in Figure 7, illustrate the accuracy comparisons of the six models across five training runs. Additionally, Table 4 summarizes the average accuracy and standard deviation of different models during the training and testing phases, where the standard deviation reflects the stability of the model results. Notably, BP, KNN, and RF models, due to their lack of temporal dependency modeling capabilities, are unsuitable for directly handling raw time-series data. To address this, a sliding window calculation method was employed to extract statistical features, including mean, variance, maximum, minimum, and norm, resulting in five features used as inputs for non-temporal models.

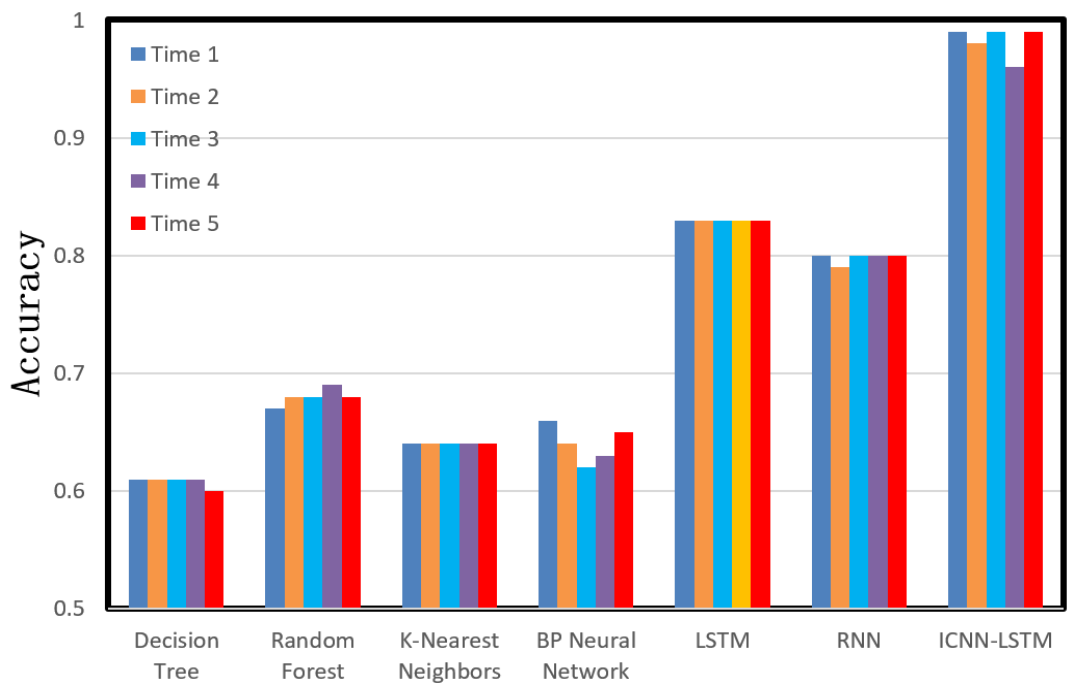

Fig. 7. Comparison of classification accuracy of dangerous state in different models

Confusion matrices were plotted for each model to observe the diagnostic accuracy across three operational conditions. The results are presented in Figure 8, where the x-axis and y-axis represent the predicted labels and the true labels, respectively. From the confusion matrices and accuracy values, it is evident that the RF, KNN, and BP models perform poorly in diagnosing the "Hazard Occurred" state, with KNN exhibiting the worst diagnostic performance. This is because, in the "Hazard Occurred" state, the number of adhered magnetic plates is minimal, resulting in low adhesion stiffness of the robot. Consequently, the vibration features amplified by the vibration rod are interfered with by the robot's own vibrations, making it difficult to extract meaningful vibration features.

In contrast, the RNN and LSTM models show a significant improvement in overall recognition accuracy. However, their accuracy remains suboptimal in the "Hazard Occurred" state. The improved ICNN-LSTM model, as discussed in this paper, demonstrates good classification performance across all three states. Both in individual state classification and overall accuracy, the ICNN-LSTM model significantly outperforms the other models.

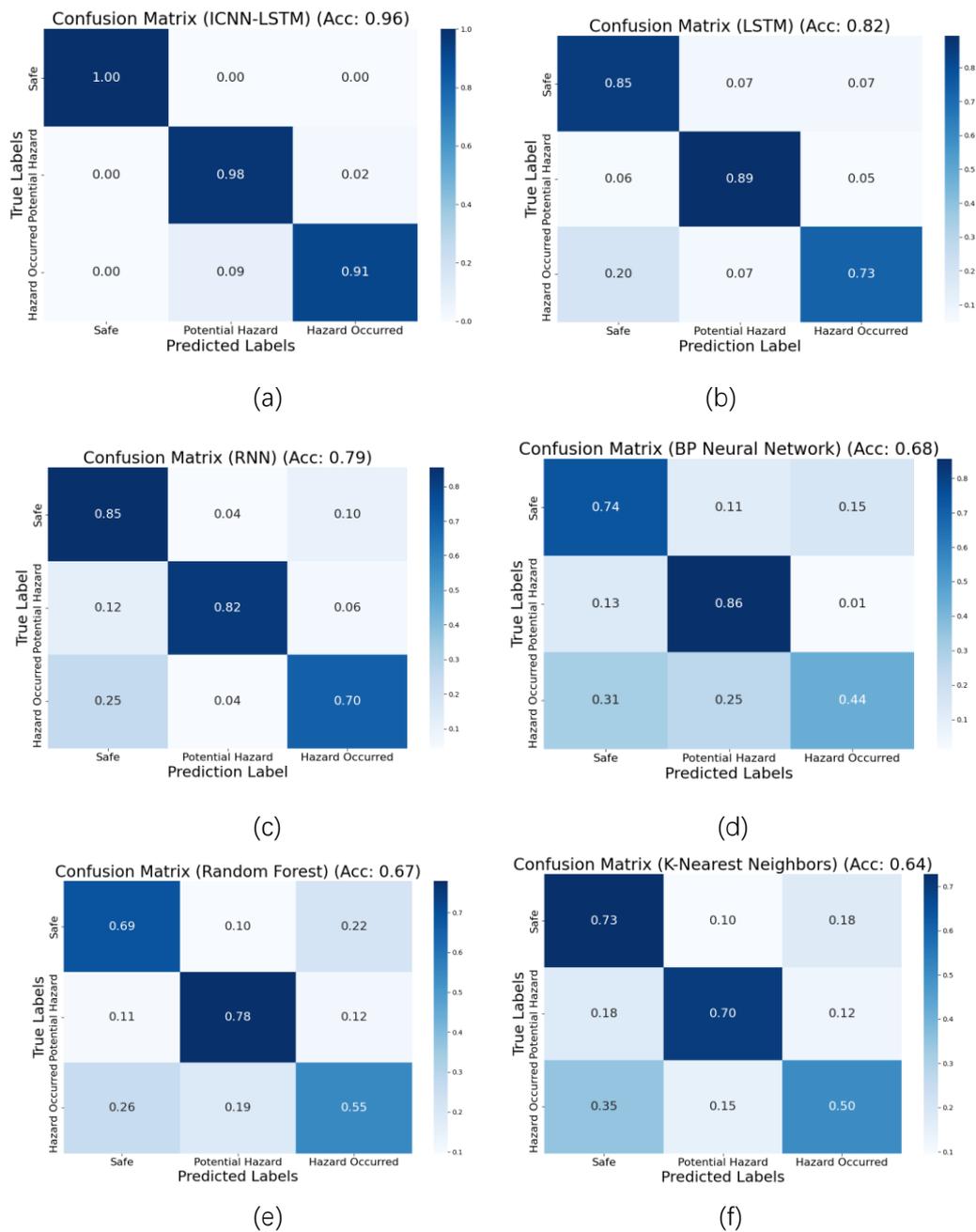

Fig. 8. Confusion matrix of classification results of different models. (a) ICNN-LSTM; (b) LSTM; (c) RNN; (d) BP; (e) RF; (f) KNN

5. Conclusion

This study focuses on enhancing the safety state perception of tracked climbing robots by designing effective and easily implementable data acquisition strategies, as well as feature extraction and classification models. Through the analysis of the climbing process of tracked climbing robots, we introduced a simple yet effective attitude data acquisition strategy that employs a carbon fiber vibration rod with an attitude sensor mounted at its end to monitor the robot's adhesion posture to the wall surface. To learn discriminative features from attitude data, we proposed an ICNN-LSTM feature extraction and classification framework, which combines an Adaptive Convolutional Neural Network (for denoising and robust feature extraction) with a Long Short-Term Memory network (for determining the number of magnetic plate attachments to the wall).

We established a testing platform comprising a climbing robot and an angle-adjustable wall surface to construct various recognition tasks. Initially, we conducted comparative experiments using two sensors. The results indicate that the proposed data acquisition strategy utilizing a carbon fiber vibration rod achieves superior performance under low to medium vibration excitation conditions, both in terms of sensor data quality and model classification results, compared to the direct data acquisition strategy on the robot body. Furthermore, comparative model evaluations demonstrated that the proposed ICNN-LSTM model significantly outperforms other popular classification models in classification accuracy.

Currently, there is no systematic design theory guiding the design and use of vibration rods. Therefore, our future research will focus on theoretically formalizing the design and utilization of vibration rods to enhance their adaptive vibration sensing capabilities. Additionally, in future studies, the proposed ICNN-LSTM model will be extended to various robotic perception classification tasks to achieve broader applicability and improved generalization across different robotic systems.